\documentclass[journal]{IEEEtran}
\usepackage{booktabs}
\usepackage{multirow}
\usepackage{threeparttable}

\usepackage{cite}

\ifCLASSINFOpdf
  \usepackage[pdftex]{graphicx}
\else
\fi
\usepackage{amsmath,amsfonts,amssymb}
\usepackage{bm}
\usepackage{times}
\usepackage{xcolor, color, colortbl}
\usepackage{float}
\usepackage{bbding}
\usepackage[capitalize]{cleveref} 
\crefname{section}{Sec.}{Secs.}
\Crefname{section}{Section}{Sections} 
\Crefname{table}{Table}{Tables}
\crefname{table}{Tab.}{Tabs.}
\newcommand{\Tref}[1]{Table~\ref{#1}}

\newcommand{\Sref}[1]{Section~\ref{#1}}
\newcommand{\eref}[1]{Eq.~(\ref{#1})}
\newcommand{\fref}[1]{Fig.~\ref{#1}}

\newcommand{\argmax}{\operatornamewithlimits{argmax}}

\renewcommand{\paragraph}[1]{\vspace{1.25mm}\noindent\textbf{#1}}
\newcommand{\method}{ILM-ASSL\xspace}
\newcommand{\ssl}{semi-supervised learning\xspace}
\newcommand{\Ssl}{Semi-supervised learning\xspace}

\newcommand{\Loss}{\mathcal{L}}
\newcommand{\Losssup}{\mathcal{L}_{s}}
\newcommand{\Lossunsup}{\mathcal{L}_{u}}
\newcommand{\Losscontrast}{\mathcal{L}_{c}}

\newcommand{\Source}{\mathcal{D}_s}
\newcommand{\Target}{\mathcal{D}_t}

\newcommand{\etal}{\textit{et al.}}
\definecolor{Gray}{gray}{0.9}

\usepackage{algorithmic}
\usepackage[linesnumbered,ruled,vlined]{algorithm2e}
\begin{document}
%
\title{Iterative Loop Method Combining Active and Semi-Supervised Learning for Domain \\Adaptive Semantic Segmentation}
%
%
%


\author{Licong Guan,
        Xue Yuan
\thanks{Corresponding author: Xue Yuan}
\thanks{The source code and pre-trained models are available at: https://github.com/licongguan/ILM-ASSL}
\thanks{The authors are with the School of Electronic and Information Engineering, Beijing Jiaotong University, Beijing, 
100044, China (e-mail: xyuan@bjtu.edu.cn).}}

\maketitle

\begin{abstract}
Semantic segmentation is an important technique for environment perception in intelligent transportation systems. With the rapid development of convolutional neural networks (CNNs), road scene analysis can usually achieve satisfactory results in the source domain. However, guaranteeing good generalization to different target domain scenarios remains a significant challenge. Recently, semi-supervised learning and active learning have been proposed to alleviate this problem. Semi-supervised learning can improve model accuracy with massive unlabeled data, but some pseudo labels containing noise would be generated with limited or imbalanced training data. And there will be suboptimal models if human guidance is absent. Active learning can select more effective data to intervene, while the model accuracy can not be improved because the massive unlabeled data are not used. And the probability of querying sub-optimal samples will increase when the domain difference is too large, increasing annotation cost. This paper proposes an iterative loop method combining active and semi-supervised learning for domain adaptive semantic segmentation. The method first uses semi-supervised to learn massive unlabeled data to improve model accuracy and provide more accurate selection models for active learning. Secondly, combined with the predictive uncertainty sample selection strategy of active learning, manual intervention is used to correct the pseudo-labels. Finally, flexible iterative loops achieve the best performance with minimal labeling cost. Extensive experiments show that our method establishes state-of-the-art performance on tasks of GTAV $\to$ Cityscapes, SYNTHIA $\to$ Cityscapes, improving by 4.9\% mIoU and 5.2\% mIoU, compared to the previous best method, respectively.
\end{abstract}


%
\IEEEpeerreviewmaketitle

\section{Introduction}
\label{sec:intro}
%
%
%
%
\IEEEPARstart{E}{nvironmental} awareness is particularly important for intelligent transportation systems (ITS), and the key to accurate environmental awareness lies in semantic segmentation~\cite{9529067, 9042876}. Benefiting from the rapid development of deep learning, many advanced segmentation methods have been proposed and achieved breakthroughs in various tasks such as autonomous driving~\cite{geiger2012we}, scene parsing~\cite{cordts2016cityscapes}, and medical analysis~\cite{asgari2021deep}. Semantic segmentation can understand image scenes at the pixel level, but it is very eager for images carefully annotated by human annotators~\cite{zhang2018context}. Especially in the above domains, massively annotated images are expensive, time-consuming~\cite{cordts2016cityscapes}, or even infeasible, which greatly hinders their widespread application. In addition, when the source domain model is applied to other target domain scenarios, due to domain differences, the generalization ability of the source domain model is poor, and it is difficult to directly promote the application~\cite{10021219}. Therefore, for ITS, how to guarantee the good generalization ability of the source domain model with the minimum annotation cost is a challenge.

\begin{figure}[!t]
  \centering
  \includegraphics[width=1.0\linewidth]{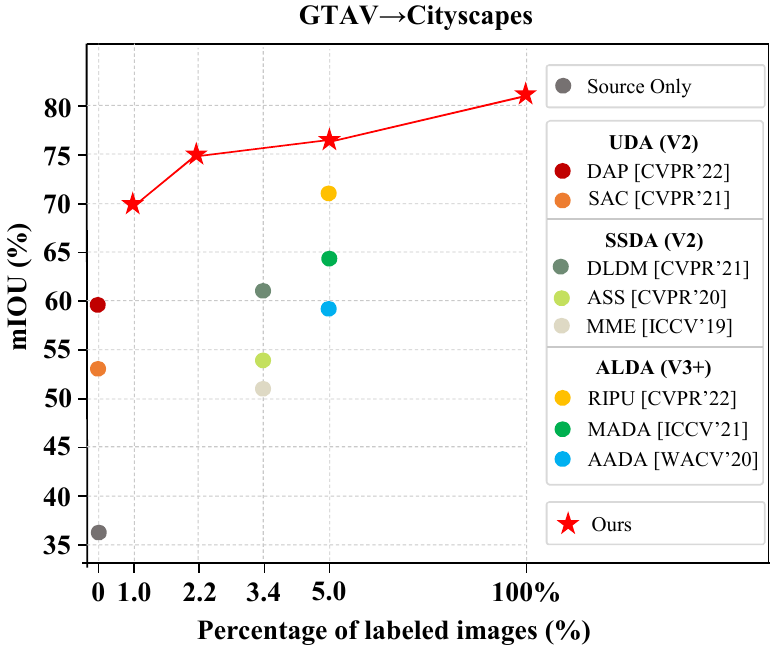}
  \caption{Performance comparison between our method and other methods on GTAV $\to$ Cityscapes. V2 and V3+ are based on DeepLab-v2 and DeepLab-V3+, respectively.}
  \label{fig1}
  \vspace{-4mm}
\end{figure}

Nowadays, many research works have proved that domain adaptation is one of the powerful means to address the above issues~\cite{li2021generalized, vu2019advent}. As shown in \fref{fig1}, in the domain adaptation task from synthetic data GTAV to real data Cityscapes, unsupervised domain adaptation (UDA), semi-supervised domain adaptation (SSDA), and active learning domain adaptation (ALDA) have achieved good performance. Among them, UDA~\cite{huo2022domain, zhang2021transfer, song2020nighttime, liu2021source, zhang2021prototypical} aims to solve this problem by leveraging the knowledge of label-rich data (source data) and transferring it to unlabeled data (target data). While it can avoid the intensive workload of manual annotation, the performance still lags far behind fully supervised models~\cite{shin2021labor}. Although SSDA improves performance by learning from unlabeled target domain data, the noise of pseudo-labels cannot be resolved. Furthermore, active learning can significantly improve performance on both classification and detection tasks by introducing a few additional manual annotations to a few selected samples from the target domain~\cite{su2020active}. Nevertheless, ALDA does not utilize the massive unlabeled data in the target domain, and only improves the accuracy through manual intervention, which makes the labeling cost difficult to control. And the probability of querying sub-optimal samples will increase when the domain difference is too large, increasing annotation cost. Some previous studies on active learning domain adaptation are based on pixel selectors~\cite{shin2021labor}. Although good results have been achieved on the Cityscapes dataset, according to existing literature and practical research, there is currently no human-computer interface for single-point pixel-level semantic annotation in real environments. In the actual manual labeling process, whether this method based on pixel selection can reduce the amount of labeling has not been verified. Therefore, how to design a practical active selection strategy based on the existing labeling software Labelme with a single image as the smallest selection unit is one of the urgent problems to be solved.

Recently, \ssl has facilitated domain adaptation, as it can retrain the network with pseudo-labels generated from massive unlabeled data~\cite{8733203, mei2020instance, zheng2021rectifying, zou2018unsupervised, zou2019confidence}. However, due to the limited and imbalanced training data, the pseudo-labels generated by \ssl usually contain noise. This disadvantageous experience not only can not improve the accuracy of pseudo-labels but also further affect the performance of machine learning models without timely manual intervention and guidance. Therefore, how discovering and correcting false pseudo-labels in the \ssl process  is one of the problems to be solved urgently.

To solve the above problems, this paper proposes \method for domain-adaptive semantic segmentation. It learns massive unlabeled data semi-supervised to improve the accuracy in the target domain and provides an accurate selection model for active learning. Then, combined with the predictive uncertainty sample selection strategy of active learning, manual intervention is used to correct the pseudo-labels. \Ssl and active learning complement each other and help the model achieve the best performance with minimal annotation cost. In a nutshell, our contributions can be summarized as: 
\begin{itemize}
  \item We propose an iterative loop method combining active learning with semi-supervised learning, termed \method. The method achieves the best performance on two representative domain adaptation benchmarks with minimal labeling cost, i.e., GTAV $\to$ Cityscapes, SYNTHIA $\to$ Cityscapes.
  \item We introduce semi-supervised learning in the active learning domain adaptation semantic segmentation task, which improves the performance of the model by learning from a large amount of unlabeled data, and corrects the semi-supervised pseudo-label noise problem with limited human intervention.
  \item From the perspective of practical application, based on the existing labeling software Labelme, we propose a selection strategy based on prediction uncertainty with a single image as the smallest selection unit. The image label is corrected in combination with the prediction of the model, which further reduces the labeling cost.
\end{itemize}

The remainder of this paper is organized as follows. In Section II, the related works of autonomous driving, domain adaptation, semi-supervised learning, and active learning are reviewed. In Section III, detailed descriptions of our approach are presented. The experimental datasets, implementation details, comparative experiments, and qualitative results are described in Section IV. Finally, section V contains some concluding remarks.

\section{Related Work}
\label{sec:related}

\subsection{Autonomous Driving}
Semantic segmentation can assist vehicles to perceive and understand the surrounding environment and is a key part of autonomous driving technology~\cite{9529067}. In recent years, speed is an important factor in the analysis of urban street scenes for intelligent transportation systems~\cite{9619854}, and real-time semantic segmentation has attracted much attention because it can generate high-quality segmentation results in real-time~\cite{9042876, 9134735}. Previous studies have shown that with the continuous expansion of the source domain training data, the model accuracy will also be significantly improved. But when it is applied to a target domain that has an inconsistent data distribution with the source domain, the accuracy of semantic segmentation will drop drastically~\cite{10021219}. Therefore, how to quickly and cost-effectively deploy the source domain model to the actual application target domain scene is the main problem faced by domain-adaptive semantic segmentation in autonomous driving technology.

\subsection{Domain Adaptation}
Domain adaptation (DA) can transfer knowledge from a label-rich source domain to a label-scarce target domain, and recent work has achieved great success on a range of tasks. Such as classification~\cite{li2021transferable}, detection~\cite{vs2021mega}, and segmentation~\cite{liu2021source}. Most previous works have used adversarial learning~\cite{vu2019advent, wang2020differential} in an attempt to reduce the domain gap between source and target features from the image level or feature level. Recent work on domain-adaptive semantic segmentation can be mainly divided into two categories: adversarial training-based methods~\cite{vu2019advent, wang2020classes} and self-training-based methods~\cite{zou2019confidence, zheng2021rectifying, zhang2021prototypical}. For the first branch, most works tend to learn domain-invariant representations based on min-max adversarial optimization games by tricking the domain discriminator to obtain aligned feature distributions~\cite{vu2019advent, wang2020classes}. To enforce local semantic consistency, Luo \etal~\cite{luo2019taking} propose to adaptively weigh the adversarial loss. The second branch focuses on how to generate high-quality pseudo-labels for target domain data for further model optimization~\cite{zheng2021rectifying, zhang2021prototypical}, which drives the development of self-training techniques~\cite{saito2019semi, chen2021semi}. In this work, we exploit the generated pseudo-labels to learn the information hidden from massive unlabeled data, thereby improving the performance of domain-adaptive semantic segmentation.

\subsection{Semi-Supervised Learning}
Semi-supervised learning (SSL) involves two paradigms: consistency regularization~\cite{ouali2020semi} and entropy minimization ~\cite{hu2021semi, wang2022semi}. Consistency regularization forces the model to produce stable and consistent predictions on the same unlabeled data under various perturbations~\cite{xie2020unsupervised}. On the other hand, entropy minimization, generalized by self-training pipelines~\cite{araslanov2021self}, exploits unlabeled target domain data in a way that uses pseudo-labels for training. For example, Wang \etal~\cite{wang2020differential} propose the Semantic-Level Shift (ASS) framework, which introduces an additional semantic-level adaptation module by adversarial training on the corresponding outputs of the source and target labeled inputs. However, adversarial loss makes training unstable due to weak supervision. Zou \etal~\cite{zou2018unsupervised} proposed an iterative learning strategy with class-balanced and spatial priors for target instances. Hu \etal~\cite{hu2021semi} proposed an adaptive balanced semi-supervised semantic segmentation framework to address the problem of unbalanced training on long-tailed data. Wang \etal~\cite{wang2022semi} exploit unreliable pixels by adding a contrastive learning loss on top of \ssl. The above studies can improve the model accuracy of the source domain by using massive unlabeled data, but they cannot fundamentally solve the problem of domain adaptation, and the noise problem of pseudo-labels cannot be effectively solved. Therefore, this paper incorporates a predictively uncertain sample selection strategy of active learning to correct the pseudo-labels during semi-supervised learning to improve the performance of domain-adaptive semantic segmentation.

\subsection{Active Learning}
Active learning (AL) aims to maximize the model performance with the least labeling cost of the dataset. Query rules are the core content of active learning, and  commonly used query strategies are divided into uncertainty-based methods~\cite{du2017robust} and diversity-based methods~\cite{gal2017deep}. In the field of active learning domain adaptation research, previous work has mainly focused on classification tasks~\cite{fu2021transferable, prabhu2021active}. Ning \etal~\cite{ning2021multi} and Shin \etal~\cite{shin2021labor} were the first to adopt active learning to solve domain-adaptive semantic segmentation tasks. Among them, \cite{ning2021multi} proposed a multi-anchor strategy to actively select image subsets, which can be inefficient. \cite{shin2021labor} proposed a more efficient point-based annotation. However, the selected points ignore the pixel spatial continuity of the image. Recently, Xie \etal~\cite{xie2022towards} greatly improved the segmentation performance in the target domain by exploring the consistency of the image space and selecting the most diverse and uncertain image regions. Although \cite{shin2021labor} and \cite{xie2022towards} achieved good results in public datasets, this pixel-based selection method did not visually verify the reduced annotation workload in the manual annotation process. The labeling software currently used for semantic segmentation is Labelme, which uses a single image as the smallest unit, and completes the labeling operation with mouse clicks based on the outline of the object. The most important thing is that the above-mentioned active learning research work ignores the utilization of massive unlabeled data in the target domain and only relies on limited labeled data for training, resulting in high labeling costs. From the perspective of practical promotion and application, this paper proposes an active selection strategy with a single image as the minimum selection unit. Moreover, this paper uses semi-supervised learning to improve the accuracy of the model and generate pseudo-labels for unlabeled data. Use the labeling software Labelme to make minor modifications to further reduce the cost of labeling and improve the performance of domain-adaptive semantic segmentation.

\section{Approach}
\label{sec:method}

\begin{figure*}[!thp]
  \centering
  \includegraphics[width=0.95\linewidth]{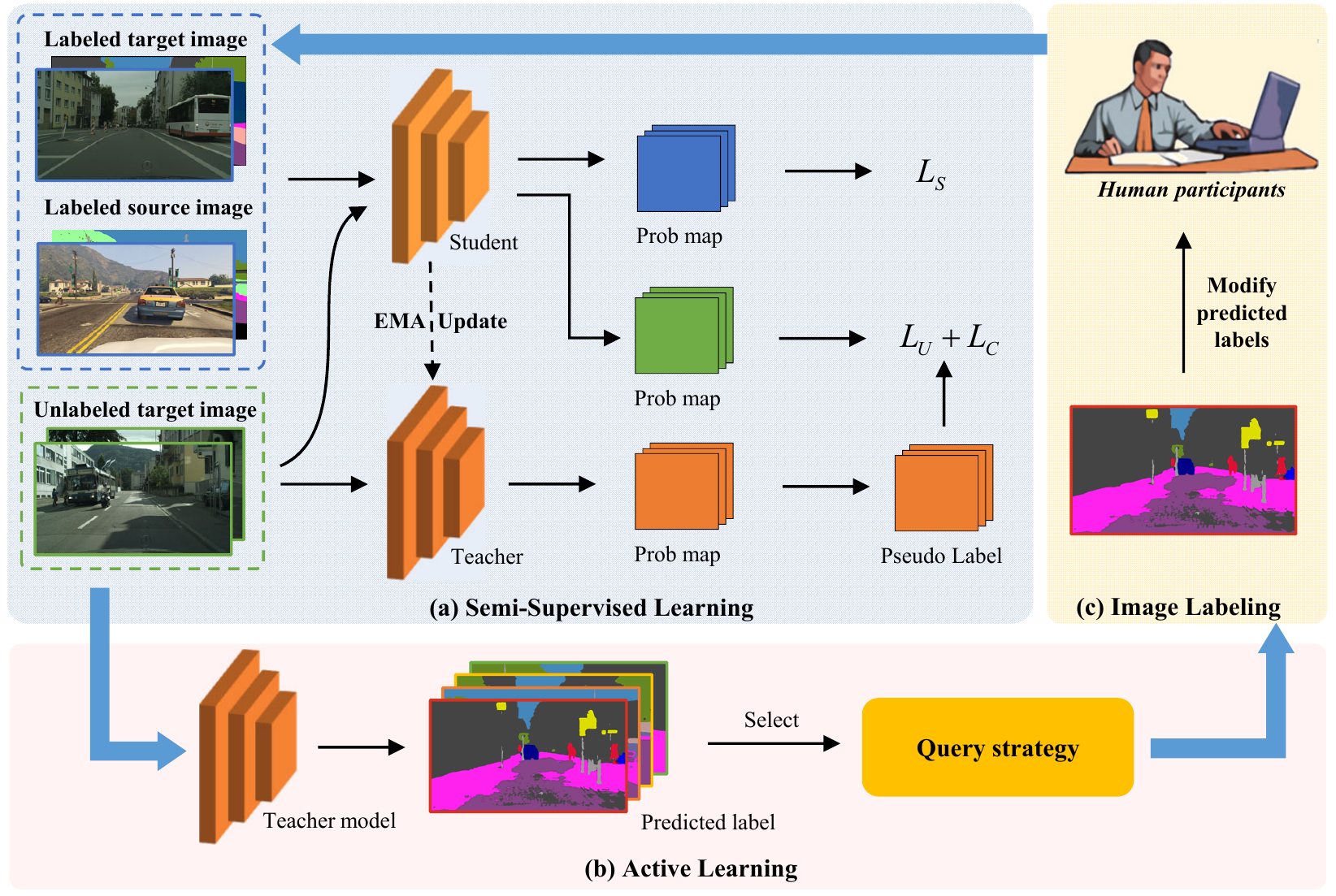}
  \caption{{\bf The overview of the proposed \method.} The proposed framework consists of three stages: the first stage (a) is \ssl. Semi-supervised learning uses a very small amount of labeled data and a large amount of unlabeled data, resulting in a teacher model. The second stage (b) is active learning. A large amount of unlabeled target domain data is predicted by the teacher model, and the prediction results are sorted according to the uncertainty selection strategy of active learning, and then the corresponding number of prediction results and corresponding original images are selected according to the labeling budget. The third stage (c) is image labeling. Experts modify the labels of unreliable prediction results, add the modified samples to the labeled target data set, and repeat the first stage of semi-supervised learning to obtain the final model.}
  \label{fig2}
  \vspace{-4mm}
\end{figure*}

In this section, we establish our problem mathematically and first outline our proposed method in \Sref{sec:overview}. Our strategies for \ssl and active learning are presented in \Sref{sec:semi-supervised learning} and \Sref{sec:active learning}, respectively.

\subsection{Overview}
\label{sec:overview}

The success of CNN-based methods benefits from a large amount of manually labeled data and the assumption of independent and identical data distributions between training and testing samples. However, when a model trained on the training set (source domain) is directly applied to an unseen test scene (target domain), the performance drops significantly. To transfer knowledge efficiently, recent advances~\cite{saito2019semi, wang2020differential, chen2021semi} employ \ssl and optimize the cross-entropy loss using target pseudo-labels. Due to limited training data and class imbalance, the pseudo-labels generated by \ssl often contain noise, and when lacking human intervention, inferior experience during training can lead to sub-optimal models. Further, active learning can pick out more effective data to intervene. However, active learning does not use massive amounts of unlabeled data, and the probability of querying sub-optimal samples will increase when the domain difference is too large, increasing annotation cost.

To solve the above problems, we propose \method. we have a set of labeled source domain data $\Source=\{(\mathbf{x}_{s}, \mathbf{y}_{s})\}$ and incompletely labeled target data $\Target=\{(\mathbf{x}_{t},\mathbf{y}_{t})\}$. where $\mathbf{y}_{s}$ is the pixel label belonging to one of the $C$ known classes in the label space $\mathcal{Y}$. Our goal is to learn a semantic segmentation network $f \circ h: \mathbf{x} \to \mathbf{y}$ parameterized by $\theta$. It leverages source domain data and a small amount of target domain labeled data to achieve good segmentation performance in both source and target domain scenarios. The proposed framework consists of three stages: the first stage is \ssl (\fref{fig2}\textcolor{red}{a}). Semi-supervised learning uses a very small amount of labeled data and a large amount of unlabeled data, resulting in a teacher model. The second stage is active learning (\fref{fig2}\textcolor{red}{b}). A large amount of unlabeled target domain data is predicted by the teacher model, and the prediction results are sorted according to the uncertainty selection strategy of active learning, and then the corresponding number of prediction results and corresponding original images are selected according to the labeling budget. The third stage is image labeling (\fref{fig2}\textcolor{red}{c}). Experts modify the labels of unreliable prediction results, add the modified samples to the labeled target data set, and repeat the first stage of semi-supervised learning to obtain the final model.

\subsection{Semi-Supervised Learning}
\label{sec:semi-supervised learning}

\method follows a typical \ssl framework, which consists of a student model and a teacher model~\cite{hoyer2022daformer}. The teacher model and the student model have the same schema. The two models differ only in updating their weights, the student's weight $\theta_{s}$ is updated by convention, while the teacher model's weight $\theta_{t}$ is updated by the exponential moving average (EMA) of the student model. For labeled images of source and target domains, we use standard cross-entropy loss on them. And for each unlabeled target domain image, we bring it into the teacher model for prediction and obtain the pseudo-label according to the pixel prediction entropy. Subsequently, the student model is trained on unlabeled target domain data and corresponding pseudo-labels. Our optimization objective is to minimize the overall loss, which can be expressed as:
\begin{equation}
  \Loss = \Losssup + \lambda_{u}\Lossunsup + \lambda_{c}\Losscontrast \,,
  \label{equ1}
\end{equation}
where $\Losssup$ (\eref{equ2}) and $\Lossunsup$ (\eref{equ3}) represent the supervised and unsupervised loss applied to labeled and unlabeled images, respectively. $\Losscontrast$ (\eref{equ4}) is the contrastive learning loss~\cite{oord2018representation}. $\lambda_{u}$ is the weight of the unsupervised loss, $\lambda_{c}$ is the weight of the contrastive loss, and $\Losssup$ and $\Lossunsup$ are both cross-entropy (CE) loss.
\begin{equation}
  \mathcal{L}_s = \frac{1}{{N}_l} \sum_{i=1}^{N_l} \frac{1}{WH} \sum_{j=1}^{WH} \ell_{ce}(f\circ h(\mathbf{x}_{i,j}^l; \theta), \mathbf{y}_{i,j}^l) \,,
  \label{equ2}
\end{equation}
\begin{equation}
  \mathcal{L}_u = \frac{1}{{N}_u} \sum_{i=1}^{N_u} \frac{1}{WH} \sum_{j=1}^{WH} \ell_{ce}(f\circ h(\mathbf{x}_{i,j}^u; \theta), \hat{\mathbf{y}}_{i,j}^u) \,,
  \label{equ3}
\end{equation}
\begin{equation}
\begin{aligned}
  \mathcal{L}_c = &- \frac{1}{C\times M}\sum_{c=0}^{C-1} \sum_{i=1}^M \\
    &\log\left[ \frac{e^{\langle \mathbf{a}_{ci}, \mathbf{a}_{ci}^{+}\rangle / \omega}}{e^{\langle \mathbf{a}_{ci}, \mathbf{a}_{ci}^{+}\rangle / \omega} + \sum_{j=1}^N e^{\langle \mathbf{a}_{ci}, \mathbf{a}_{cij}^{-}\rangle / \omega}} \right] \,.
  \label{equ4}
\end{aligned}
\end{equation}

In \eref{equ2} and \eref{equ3}, $f\circ h$ is the composition function of $h$ and $f$, which means that the image $\mathbf{x}_{i,j}$ is first sent to $h$ to extract features, and then sent to $f$ to obtain segmentation results. $\mathbf{y}_{i,j}^l$ is the manually annotated mask label for the $j$-th pixel in the $i$-th labeled image, $\hat{\mathbf{y}}_{i,j}^u$ is the pseudo-label for the $j$-th pixel in the $i$-th unlabeled image. $N_l$ and $N_u$ represent the number of labeled and unlabeled images in the training batch. $W$ and $H$ are the width and height of the input image. $\ell_{ce}$ is the standard cross-entropy loss. In \eref{equ4}, $M$ is the total number of anchor pixels, $N$ represents the number of negative pixels used for contrastive loss training in mini-batch, and $C$ is the number of classes. $\mathbf{a}_{ci}$ represents the representation of the $i$-th anchor of class $c$. The positive and negative samples corresponding to each anchor pixel are denoted as $\mathbf{a}_{ci}^{+}$ and $\mathbf{a}_{ci}^{-}$, $\langle\cdot, \cdot\rangle$ is the cosine similarity between features from two different pixels, whose range is limited between -1 to 1 according to $\omega$.

We generate pseudo-labels $\hat{y}_{ij}^u$ by taking the entropy of the probability distribution of each pixel predicted by the teacher model. The formula for calculating entropy is as follows:
\begin{equation}
    {\mathcal{H}(\mathbf{p}_{ij}) = -\sum_{c=0}^{C-1} p_{ij}(c)\log p_{ij}(c),}
  \label{equ5}
\end{equation}
where $p_{ij}$ denotes the softmax probability generated by the teacher model at the $j$-th pixel of the $i$-th unlabeled image. $C$ is the number of classes. $c$ represents the $c$-th dimension.

During training, the accuracy of the pseudo-labels increases along with the performance of the teacher model. Therefore, we define pseudo-labels in terms of training epochs $t$ as follows:
\begin{equation}
    {\hat{y}_{ij}^u = \argmax_c p_{ij}(c), \mathrm{if}\ \mathcal{H}(\mathbf{p}_{ij}) < \gamma_t,}
  \label{equ6}
\end{equation}
where $\gamma_t$ represents the entropy threshold of the $t$-th training step, and we set $\gamma_t$ as the quantile corresponding to $\alpha_t$, that is: $\gamma_t$=\texttt{np.percentile($\mathcal{H}$.flatten(),100*(1-$\alpha_t$))}, where $\mathcal{H}$ is per-pixel entropy map.

We adjust at each epoch with a linear strategy as follows:
\begin{equation}
    {\alpha_t = \alpha_0 \cdot \left(1 - \frac{t}{\mathrm{total\ epoch}}\right),}
  \label{equ7}
\end{equation}
where $\alpha_0$ is the initial value, which we set as 20\%, and $t$ is the current training epoch.

\subsection{Active Learning}
\label{sec:active learning}

In \Sref{sec:semi-supervised learning}, the teacher model can be used to predict the unlabeled data to generate pseudo-labels for unsupervised training. However, when the distributions of the source domain and the target domain are quite different, and the labeled data of the target domain is limited, the model cannot learn enough features of the target domain, thus generating unreliable pseudo-labels. These unreliable pseudo-labels can further affect the performance of machine-learning models if there is no timely and effective human intervention. Therefore, based on semi-supervised learning, this paper uses the teacher model to predict a large number of unlabeled target domain data and ranks the prediction results according to the uncertainty selection strategy of active learning. Experts modify the labels of unreliable prediction results, add the modified samples to the labeled target data set, and semi-supervised training is performed again to complete an iteration. In practical applications, the maximum performance of domain-adaptive semantic segmentation can be achieved with minimal human intervention cost through multiple iterative loops according to the annotation budget.

Our proposed query strategy based on prediction uncertainty is as follows: 

(1) Given an unlabeled target image $\mathbf{x}_{i,j}^u$ and the teacher model $\theta_{t}$ obtained in \Sref{sec:semi-supervised learning}. According to \eref{equ5}, the teacher model is used to predict the unlabeled image, and the entropy of the probability distribution of each pixel in the image is obtained. 

(2) To calculate the uncertainty score for each image, we take the average of all entropies as the uncertainty score for that image. If the uncertainty score of the image is high, it means that the model predicts the image poorly. At the same time, in the process of semi-supervised learning, the noise of the pseudo-label generated by the image is large, which directly affects the performance of the model. Therefore, we artificially intervene in these poorly predicted results to correct wrong pseudo-labels in time. The uncertainty score of an image is calculated as follows:
\begin{equation}
  \bm{S} = - \frac{1}{WH}\sum_{j=1}^{WH} \sum_{c=0}^{C-1} p_{ij}(c)\log p_{ij}(c),
  \label{equ8}
\end{equation}
where $W$ and $H$ are the width and height of the feature map, respectively, $p_{ij}$ represents the prediction result of the $c$-th channel of the $j$-th pixel in the $i$-th image, and $C$ represents the number of classes.

(3) We rank the prediction results of all unlabeled images from high to low by the uncertainty score to generate a sample library. According to the labeling budget, we sequentially select the corresponding number of prediction results and corresponding original images from the sample library and send them to experts for label correction. The modified data is added to the labeled target domain dataset, and semi-supervised training is performed again to obtain the final model.

\section{Experiments}
\label{sec:experiment}

\begin{table*}[t]
  \centering
  \caption{\textbf{Comparison with previous results on task GTAV $\to$ Cityscapes.} We report the mIoU and the best results are shown in \textbf{bold}.} \label{tab1}
  \vspace{-2mm}
  \resizebox{\textwidth}{!}{
  \begin{threeparttable}
  \begin{tabular}{l c c c c c c c c c c c c c c c c c c c c c}
  \toprule[1.2pt]
  Method & Net. & \rotatebox{60}{road} & \rotatebox{60}{side.} & \rotatebox{60}{buil.} & \rotatebox{60}{wall} & \rotatebox{60}{fence} & \rotatebox{60}{pole} & \rotatebox{60}{light} & \rotatebox{60}{sign} & \rotatebox{60}{veg.} & \rotatebox{60}{terr.} & \rotatebox{60}{sky} & \rotatebox{60}{pers.} & \rotatebox{60}{rider} & \rotatebox{60}{car} & \rotatebox{60}{truck} & \rotatebox{60}{bus} & \rotatebox{60}{train} & \rotatebox{60}{motor} & \rotatebox{60}{bike} & mIoU\\
  \midrule
  Source Only &\multirow{7}[1]{*}{V2} & 75.8 & 16.8 & 77.2 & 12.5 & 21.0 & 25.5 & 30.1 & 20.1 & 81.3 & 24.6 & 70.3 & 53.8 & 26.4 & 49.9 & 17.2 & 25.9 & 6.5 & 25.3 & 36.0 & 36.6 \\
  CBST~\cite{zou2018unsupervised} & & 91.8 & 53.5 & 80.5 & 32.7 & 21.0 & 34.0 & 28.9 & 20.4 & 83.9 & 34.2 & 80.9 & 53.1 & 24.0 & 82.7 & 30.3 & 35.9 & 16.0 & 25.9 & 42.8 & 45.9 \\
  MRKLD~\cite{zou2019confidence} & & 91.0 & 55.4 & 80.0 & 33.7 & 21.4 & 37.3 & 32.9 & 24.5 & 85.0 & 34.1 & 80.8 & 57.7 & 24.6 & 84.1 & 27.8 & 30.1 & 26.9 & 26.0 & 42.3 & 47.1 \\
  ASS~\cite{wang2020differential} & & 90.6 & 44.7 & 84.8 & 34.3 & 28.7 & 31.6 & 35.0 & 37.6 & 84.7 & 43.3 & 85.3 & 57.0 & 31.5 & 83.8 & 42.6 & 48.5 & 1.9 & 30.4 & 39.0 & 49.2 \\
  SAC~\cite{araslanov2021self} & & 90.4 & 53.9 & 86.6 & 42.4 & 27.3 & 45.1 & 48.5 & 42.7 & 87.4 & 40.1 & 86.1 & 67.5 & 29.7 & 88.5 & 49.1 & 54.6 & 9.8 & 26.6 & 45.3 & 53.8 \\
  ProDA~\cite{zhang2021prototypical} & & 87.8 & 56.0 & 79.7 & 46.3 & 44.8 & 45.6 & 53.5 & 53.5 & 88.6 & 45.2 & 82.1 & 70.7 & 39.2 & 88.8 & 45.5 & 59.4 & 1.0 & 48.9 & 56.4 & 57.5 \\
  DAP+ProDA~\cite{huo2022domain}    & & 94.5 & 63.1 & 89.1 & 29.8 & 47.5 & 50.4 & 56.7 & 58.7 & 89.5 & 50.2 & 87.0 & 73.6 & 38.6 & 91.3 & 50.2 & 52.9 & 0.0 & 50.2 & 63.5 & 59.8 \\
  \midrule
  LabOR (2.2\%)~\cite{shin2021labor} &\multirow{3}[1]{*}{V2} & \bf 96.6 & \bf77.0 & 89.6 & 47.8 & 50.7 & 48.0 & 56.6 & 63.5 & 89.5 & \bf 57.8 & 91.6 & 72.0 & 47.3 & 91.7 & 62.1 & 61.9 & 48.9 & 47.9 & 65.3 & 66.6 \\
  RIPU (2.2\%)~\cite{xie2022towards} & &96.5 &74.1 & 89.7 & \bf 53.1 & 51.0 & 43.8 & 53.4 & 62.2 & 90.0 & 57.6 & 92.6 & 73.0 & 53.0 & 92.8 & 73.8 & 78.5 & 62.0 & 55.6 & 70.0 & 69.6 \\ 
  \bf Ours (2.2\%)   & & 96.4 & 74.6 & \bf 91.1 & 45.9 & \bf 52.4 & \bf 59.4 & \bf 67.9 & \bf 68.3 & \bf 91.4 & 50.0 & \bf 92.8 & \bf 76.2 & \bf 57.2 & \bf 93.6 & \bf 78.2 & \bf 81.3 & \bf 69.5 & \bf 58.4 & \bf 72.1 & \bf 72.5 \\
  \midrule
  \midrule
  AADA (5\%)~\cite{su2020active}  &\multirow{6}[1]{*}{V3+} &92.2 & 59.9 & 87.3 & 36.4 & 45.7 & 46.1 & 50.6 & 59.5 & 88.3 & 44.0 & 90.2 & 69.7 & 38.2 & 90.0 & 55.3 & 45.1 & 32.0 & 32.6 & 62.9 & 59.3 \\
  MADA (5\%)~\cite{ning2021multi} & & 95.1 & 69.8 & 88.5 & 43.3 & 48.7 & 45.7 & 53.3 & 59.2 & 89.1 & 46.7 & 91.5 & 73.9 & 50.1 & 91.2 & 60.6 & 56.9 & 48.4 & 51.6 & 68.7 & 64.9 \\
  RIPU (5\%)~\cite{xie2022towards} & & 97.0 & 77.3 & 90.4 &\bf54.6 & 53.2 & 47.7 & 55.9 & 64.1 & 90.2 & \bf 59.2 & 93.2 & 75.0 & 54.8 & 92.7 & 73.0 & 79.7 & 68.9 & 55.5 & 70.3 & 71.2 \\
  \bf Ours (1\%) & & 95.2 & 67.0 & 90.9 & 47.4 & 49.6 & 60.9 & 68.2 & 67.5 & 90.9 & 44.6 & 91.5 & 81.3 & 60.5 & 93.9 & 67.2 & 76.6 & 47.9 & 54.7 & 74.8 & 70.0 \\
  \bf Ours (2.2\%)   & & 96.5 & 75.6 & 91.2 & 46.7 & 53.6 & 62.1 & 70.3 & 76.0 & 91.4 & 52.1 & 94.1 & 82.0 & 60.8 & 94.4 & \bf 83.1 & 86.4 & 71.9 & 61.2 & 75.8 & 75.0 \\
  \bf Ours (5\%)  & & \bf 96.9& \bf 77.8 & \bf 91.6 & 46.7 & \bf 56.0 & \bf 63.2 & \bf 70.8 & \bf 77.4 & \bf 91.9 & 54.9 & \bf 94.5 & \bf 82.3 & \bf 61.2 & \bf 94.9 & 79.3 & \bf 88.1 & \bf 75.3 & \bf 65.8 & \bf 77.6 & \bf 76.1 \\
  \bottomrule[1.2pt]
  \end{tabular}
  \begin{tablenotes}
      \item Methods with V2 are based on DeepLab-v2~\cite{chen2017deeplab} and methods with V3+ are based on DeepLab-v3+~\cite{chen2018encoder} for a fair comparison.
    \end{tablenotes}
  \end{threeparttable}
  }
  \vspace{-3mm}
\end{table*}

\begin{table*}[t]
  \centering
  \caption{\textbf{Comparisons with previous results on task SYNTHIA $\to$ Cityscapes.} We report the mIoUs in terms of 13 classes (excluding the ``wall", ``fence", and ``pole") and 16 classes. The best results are shown in \textbf{bold}.} \label{tab2}
  \vspace{-2mm}
  \resizebox{\textwidth}{!}{
  \begin{threeparttable}
  \begin{tabular}{l l c c c c c c c c c c c c c c c c c c }
  \toprule[1.2pt]
  Method & Net. & \rotatebox{60}{road} & \rotatebox{60}{side.} & \rotatebox{60}{buil.} & \rotatebox{60}{wall*} & \rotatebox{60}{fence*} & \rotatebox{60}{pole*} & \rotatebox{60}{light} & \rotatebox{60}{sign} & \rotatebox{60}{veg.}  & \rotatebox{60}{sky} & \rotatebox{60}{pers.} & \rotatebox{60}{rider} & \rotatebox{60}{car}  & \rotatebox{60}{bus}  & \rotatebox{60}{motor} & \rotatebox{60}{bike} & mIoU & mIoU*  \\
  \midrule
  Source Only &\multirow{7}[1]{*}{V2} & 55.6 & 23.8 & 74.6 & 9.2 & 0.2 & 24.4 & 6.1 & 12.1 & 74.8 & 79.0 & 55.3 & 19.1 & 39.6 & 23.3 & 13.7 & 25.0 & 33.5 & 38.6 \\ 
  CBST~\cite{zou2018unsupervised} & & 68.0 & 29.9 & 76.3 & 10.8 & 1.4 & 33.9 & 22.8 & 29.5 & 77.6 & 78.3 & 60.6 & 28.3 & 81.6 & 23.5 & 18.8 & 39.8 & 42.6 & 48.9 \\
  MRKLD~\cite{zou2019confidence} & & 67.7 & 32.2 & 73.9 & 10.7 & 1.6 & 37.4 & 22.2 & 31.2 & 80.8 & 80.5 & 60.8 & 29.1 & 82.8 & 25.0 & 19.4 & 45.3 & 43.8 & 50.1 \\
  ASS~\cite{wang2020differential} & & 83.0 & 44.0 & 80.3 & - &- & - & 17.1 & 15.8 & 80.5 & 81.8 & 59.9 & 33.1 & 70.2 & 37.3 & 28.5 & 45.8 & - & 52.1 \\
  SAC~\cite{araslanov2021self} & & 89.3 & 47.2 & 85.5 & 26.5 & 1.3 & 43.0 & 45.5 & 32.0 & 87.1 & 89.3 & 63.6 & 25.4 & 86.9 & 35.6 & 30.4 & 53.0 & 52.6 & 59.3 \\
  ProDA~\cite{zhang2021prototypical} & & 87.8 & 45.7 & 84.6 & 37.1 & 0.6 & 44.0 & 54.6 & 37.0 & 88.1 & 84.4 & 74.2 & 24.3 & 88.2 & 51.1 & 40.5 & 45.6 & 55.5 & 62.0 \\ 
  DAP+ProDA~\cite{huo2022domain}    & & 84.2 & 46.5 & 82.5 & 35.1 & 0.2 & 46.7 & 53.6 & 45.7 & 89.3 & 87.5 & 75.7 & 34.6 & 91.7 & 73.5 & 49.4 & 60.5 & 59.8 & 64.3 \\
  \midrule
  RIPU (2.2\%)~\cite{xie2022towards} &\multirow{2}[1]{*}{V2} &96.8 &\bf76.6 &89.6 &45.0 &47.7 &45.0 &53.0 &62.5 & 90.6 & 92.7 & 73.0 & 52.9 & 93.1 & \bf80.5 & 52.4 & 70.1 & 70.1 & 75.7 \\
  \bf Ours (2.2\%)                  & & \bf 96.8 & 76.1 &\bf 89.7 & \bf47.3 & \bf52.8 & \bf56.3 & \bf62.9 & \bf70.1 & \bf91.1 & \bf93.2 & \bf78.4 & \bf59.7 & \bf93.5  & 78.2 & \bf58.2 & \bf74.2 & \bf73.7 & \bf78.6 \\  
  \midrule
  \midrule
  AADA (5\%)~\cite{su2020active}  &\multirow{6}[1]{*}{V3+}  &91.3 &57.6 &86.9 &37.6 &48.3 &45.0 &50.4 &58.5 &88.2 &90.3 &69.4 &37.9 &89.9 &44.5 &32.8 &62.5 &61.9 & 66.2 \\
  MADA (5\%)~\cite{ning2021multi}   & & 96.5 & 74.6 & 88.8 & 45.9 & 43.8 & 46.7 & 52.4 & 60.5 & 89.7 & 92.2 & 74.1 & 51.2 & 90.9 & 60.3 & 52.4 & 69.4 & 68.1 & 73.3  \\
  RIPU (5\%)~\cite{xie2022towards}  & & 97.0 & 78.9 & 89.9 & \bf 47.2 & 50.7 & 48.5 & 55.2 & 63.9 & 91.1 & 93.0 & 74.4 & 54.1 & 92.9 & 79.9 & 55.3 & 71.0 & 71.4 & 76.7 \\
  \bf Ours (1\%)                    & & 96.8 & 74.8 & 90.0 & 34.0 & 46.3 & 60.9 & 68.0 & 74.8 & 90.2 & 92.5 & 81.1 & 58.2 & 93.0 & 72.3 & 63.4 & 75.6 & 73.2 & 79.3 \\
  \bf Ours (2.2\%)                  & & 96.8 & 76.3 & 90.9 & 48.1 & 54.2 & 62.4 & 69.0 & 77.3 & 91.0 & 93.7 & 82.2 & \bf 60.3 & 94.2 & 80.0 & 63.8 & 76.0 & 76.0 & 80.9 \\
  \bf Ours (5\%)                    & &\bf97.4 &\bf80.1 & \bf 91.8 & 38.6 & \bf 55.2 & \bf 64.1 & \bf 70.9 & \bf 78.7 & \bf 91.6 & \bf 94.5 & \bf 82.7 & 60.1 & \bf 94.4 & \bf 81.7 & \bf 66.8 & \bf 77.2 & \bf 76.6 & \bf 82.1 \\
  \bottomrule[1.2pt]
  \end{tabular}
  \begin{tablenotes}
      \item Methods with V2 are based on DeepLab-v2~\cite{chen2017deeplab} and methods with V3+ are based on DeepLab-v3+~\cite{chen2018encoder} for a fair comparison.
    \end{tablenotes}
  \end{threeparttable}
  } 
  \vspace{-3mm}
  \end{table*}

\subsection{Dataset}
To verify the effectiveness of the proposed method, we evaluate our method on two popular scenarios, transferring information from synthetic images GTAV~\cite{richter2016playing} and SYNTHIA~\cite{ros2016synthia} to the real domain, the Cityscapes~\cite{cordts2016cityscapes} dataset. \textbf{GTAV} is a synthetic image dataset containing 24,966 1914$\times$1052 images, sharing 19 classes as Cityscapes. \textbf{SYNTHIA} is a synthetic urban scene dataset containing 9,400 1280$\times$760 images, sharing 16 classes as Cityscapes. \textbf{Cityscapes} is an autonomous driving dataset of real urban scenes, containing 2,975 training images and 500 validation images, each with a resolution of 2048$\times$1024. 

\subsection{Implementation Details}
All experiments are performed on NVIDIA A100 GPU with Pytorch. We adopt DeepLabv2~\cite{chen2017deeplab} and DeepLab-v3+~\cite{chen2018encoder} architectures with ResNet-101~\cite{he2016deep} pre-trained on ImageNet~\cite{deng2009imagenet} as the backbone. Regarding the training, we use the SGD optimizer with an initial learning rate of 0.0025, weight decay of 0.0001, and momentum of 0.9. For all experiments, we train about 100K iterations with a batch size of 12, and data are resized into 769$\times$769.

\paragraph{Evaluation metric.} As a common practice~\cite{mei2020instance, ning2021multi, shin2021labor, xie2022towards}, we apply slide window evaluation. We report the mean Intersection-over-Union (mIoU)~\cite{everingham2015pascal} on the Cityscapes validation set. Specifically, we report the mIoU on the shared 19 classes for GTAV $\to$ Cityscapes and report the results on 13 (mIoU*) and 16 (mIoU) common classes for SYNTHIA $\to$ Cityscapes. We also add the 19-class evaluation for the SYNTHIA $\to$ Cityscapes task in \Sref{sec:analysis_results}, and we believe that the extremely small amount of target domain data is sufficient to optimize the three classes missing from SYNTHIA.

\paragraph{Annotation budget.} Previous active learning-based selection strategies use pixels or regions as selection units, and they select a fixed proportion (2.2\% or 5\%) of each sample in the dataset. This pixel-based selection is difficult to intuitively verify the reduction of labeling workload in the actual manual labeling process, and there is no such interactive labeling software at this stage, so it is difficult to use in practical applications. According to the labeling specifications of the existing semantic segmentation labeling software Labelme, we use a single image as the smallest unit for sample selection. For a fair comparison, we only select the corresponding percentage of images. The selection process is divided into two rounds, in the first round we randomly select 1\% (30 images) from the target domain dataset for \ssl. In the second round, we selected 1.2\% (35 images) or 4\% (120 images). Therefore, we label 2.2\% or 5\% of the target domain data in total for \ssl. It should be noted that although the number of selected samples is equal, we do not label the pixels of each object in the actual labeling process, but label based on the outline of the object. Therefore, the workload of actual manual labeling is much smaller than that of the compared method.

\begin{figure*}[!htbp]
  \centering
  \includegraphics[width=0.95\linewidth]{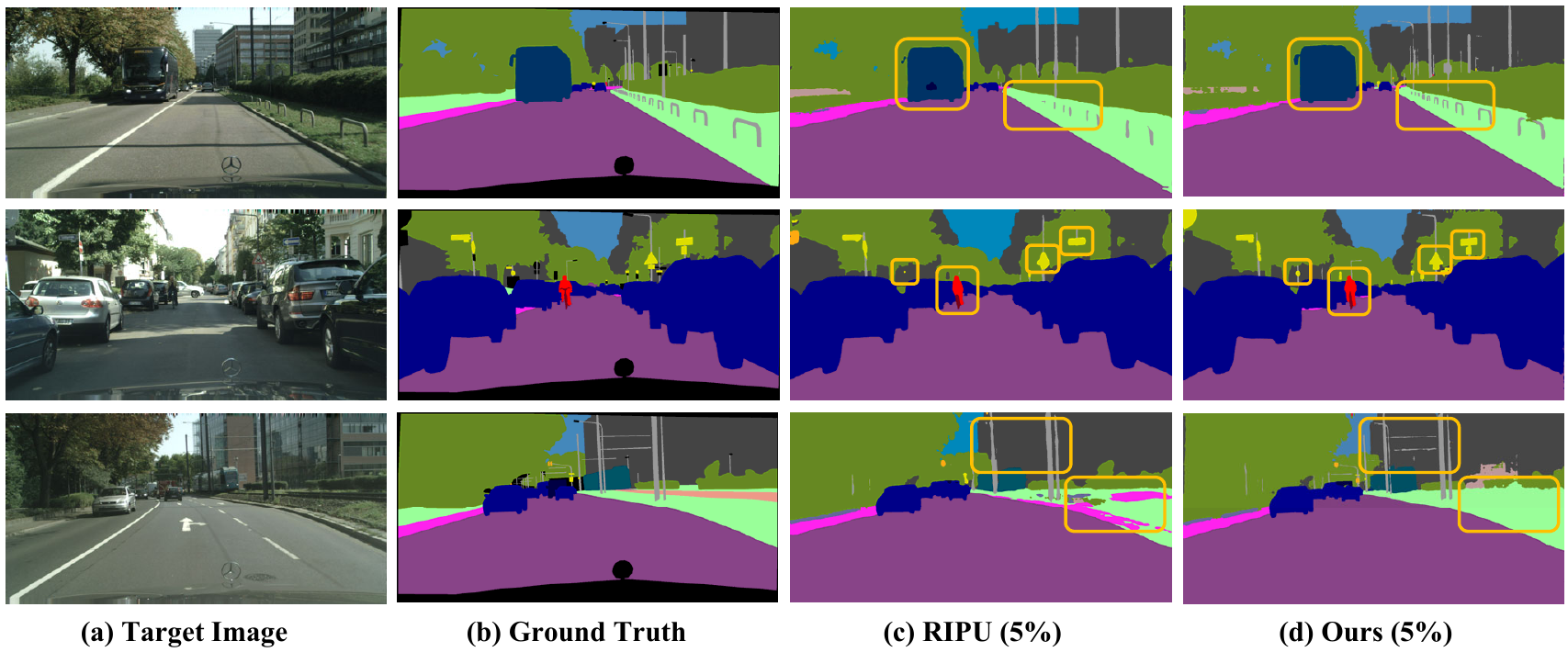}
  \caption{\textbf{Visualization of segmentation results for the task GTAV $\to$ Cityscapes.} From left to right: original target image, ground-truth label, the result predicted by RIPU~\cite{xie2022towards}, and result predicted by Ours are shown one by one.}
  \label{fig3}
  \vspace{-4mm}
\end{figure*}

\begin{figure*}[!htbp]
  \centering
  \includegraphics[width=0.95\linewidth]{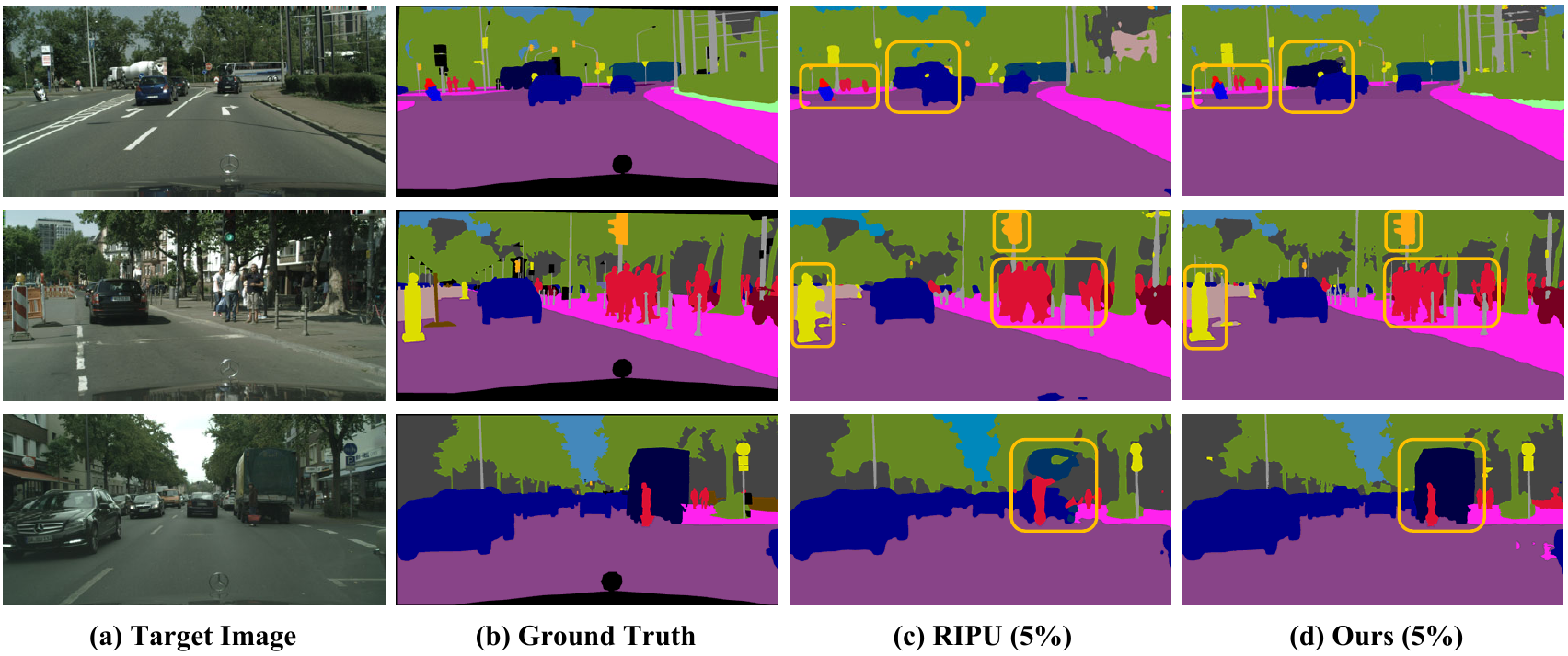}
  \caption{\textbf{Visualization of segmentation results for the task SYNTHIA $\to$ Cityscapes.} From left to right: original target image, ground-truth label, the result predicted by RIPU~\cite{xie2022towards}, and result predicted by Ours are shown one by one.}
  \label{fig4}
  \vspace{-4mm}
\end{figure*}

\begin{figure*}[!htp]
  \centering
  \includegraphics[width=0.95\linewidth]{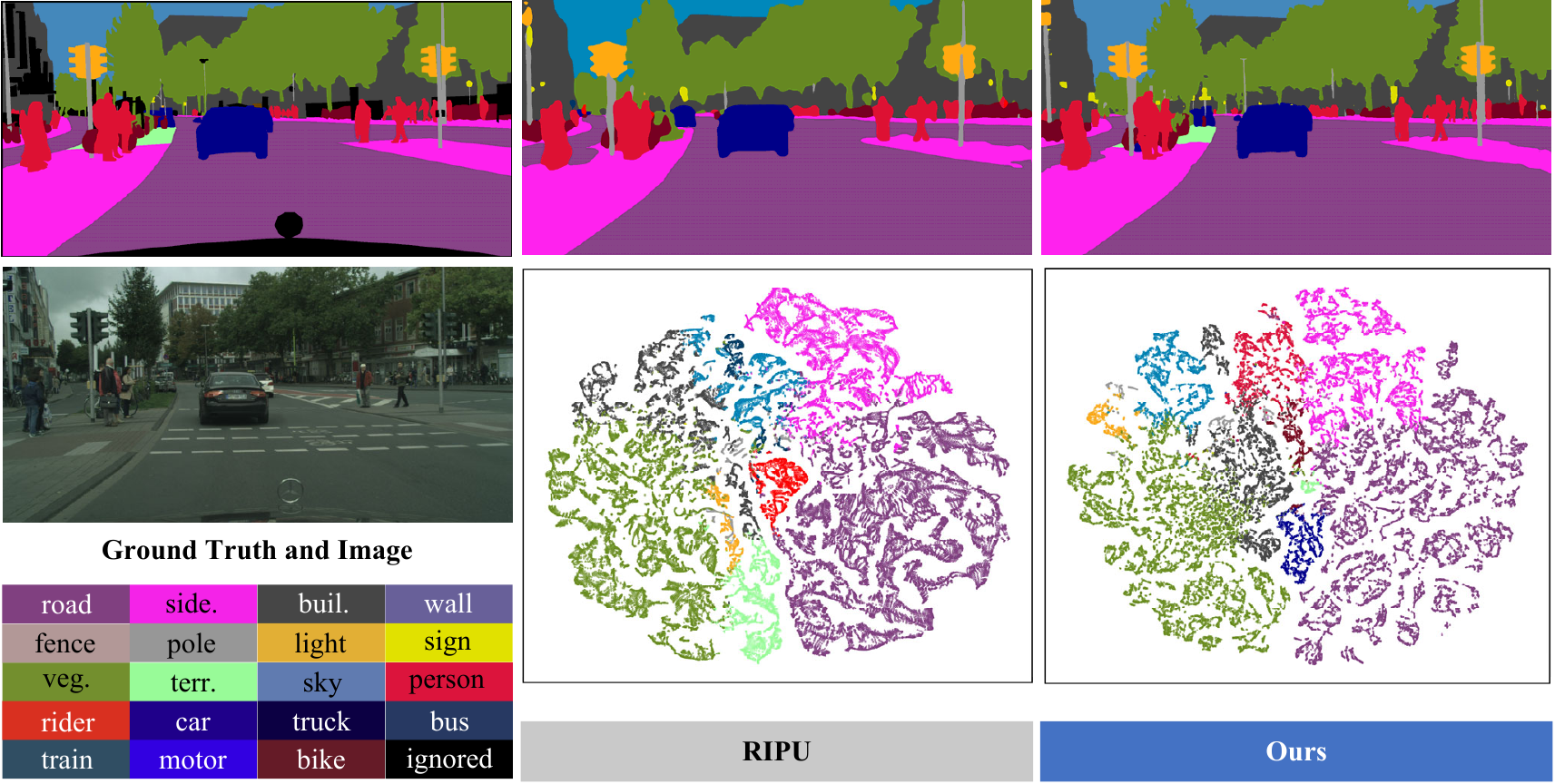}
  \caption{t-SNE analysis~\cite{van2008visualizing} of active learning method RIPU~\cite{xie2022towards} and our method. The visualization of embedded features further demonstrates that our method can exhibit the clearest clustering.}
  \label{fig5}
  \vspace{-4mm}
\end{figure*}

\subsection{Comparisons With the State-of-the-arts}
\label{sec:comparisons with the SOTA}

\Tref{tab1} and \Tref{tab2} are the domain adaptation results of GTAV $\to$ Cityscapes and SYNTHIA $\to$ Cityscapes, respectively, and it can be seen that our method greatly outperforms the previous leading unsupervised domain adaptation (ProDA~\cite{zhang2021prototypical}, DAP+ProDA~\cite{huo2022domain}) and active learning domain adaptation methods (LabOR~\cite{shin2021labor}, RIPU~\cite{xie2022towards}). At the limit, our results using only 1\% of the data are also substantial improvements over previous state-of-the-art unsupervised methods (DAP+ProDA~\cite{huo2022domain}).

For the GTAV $\to$ Cityscapes task, based on using the same backbone (DeepLab-v3+), we can easily beat AADA~\cite{su2020active} and MADA~\cite{ning2021multi} with an annotation budget of 1\%. Compared to the state-of-the-art model, using the same annotation budget (5\%), our method achieves 4.9\% mIoU improvement over RIPU~\cite{xie2022towards}. Notably, our method significantly outperforms the contrastive methods in some specific categories, namely the tail category of Cityscapes (such as ``traffic light'', ``traffic sign'', ``rider'', ``bus'', ``train'', ``motorcycle'', and ``bicycle''), which indicates that the proposed method can effectively alleviate the long-tailed distribution problem to outperform the adversary.

Our \method is still competitive for the SYNTHIA $\to$ Cityscapes task. Based on using the same backbone (DeepLab-v3+), our method can beat all methods using 1\% of the target data. Compared to the state-of-the-art model, our method achieves 5.2\% mIoU improvement over RIPU~\cite{xie2022towards} if the same annotation budget (5\%) is used. Likewise, our method outperforms RIPU~\cite{xie2022towards} on the tail categories of Cityscapes (such as ``traffic light'', ``traffic sign'', ``rider'', ``bus'', ``motorcycle'', and ``bicycle'').

\begin{table}[t]
  \centering
  \caption{Ablation study on the effectiveness of various components.}
  \label{tab3}
  \vspace{-2mm}
  \resizebox{\linewidth}{!}{
  \begin{tabular}{c c c | c}
  \toprule
  \multicolumn{2}{c}{Semi-Supervised Learning}   & Active Learning        & GT$\to$Ci \\
  \midrule
  $\Lossunsup$  & $\Losscontrast$                & Prediction uncertainty & mIoU  \\
  \midrule
                &                                &                        & 59.33 \\
  $\checkmark$  &                                &                        & 69.71  \\
  $\checkmark$  & $\checkmark$                   &                        & 70.61  \\
  $\checkmark$  & $\checkmark$                   & $\checkmark$           & \bf 76.11 \\
  \bottomrule
  \end{tabular} 
  }
  \vspace{-3mm}
\end{table}

\subsection{Qualitative Results}
\label{sec:qualitative_results}

We visualize the segmentation results predicted by \method in GTAV $\to$ Cityscapes and compare them with the state-of-the-art RIPU~\cite{xie2022towards} model prediction results. As can be seen from \fref{fig3}, our method has smoother prediction results for the head category (such as ``pole''), and for Cityscapes tail categories (such as ``bus'', ``rider'', and ``traffic sign''). The visualization of the segmentation results predicted by SYNTHIA $\to$ Cityscapes is shown in \fref{fig4}. Our method achieves accurate predictions for the missing class ``truck'' with only a small amount of target data. For the head class ``person'', \method can achieve more detailed contour prediction than adversaries. For the tail categories of Cityscapes (such as ``traffic sign'' and ``traffic light''), our predictions are greatly improved. 

\begin{table}[t]
  \centering
  \caption{Experiments with \textbf{Open Set Domain Adaptation}.} \label{tab4}
  \vspace{-2mm}
  \resizebox{\linewidth}{!}{
  \begin{tabular}{l c c c c c}
  \toprule
  Method & Net. & terrain & truck & train & mIoU\\
  \midrule
  Ours (1\%)    &\multirow{2}{*}{V3+}  & 41.5 & 51.6 & 48.6 & 69.1 \\
  Ours (2.2\%)  & & 53.5 & 74.7 & 59.9 & 73.9  \\
  \bottomrule
  \end{tabular}
  } 
  \vspace{-3mm}
  \end{table}

\begin{table}[t]
  \centering
  \caption{Comparisons with the \textbf{SSDA and SSL} methods on task GTAV $\to$ Cityscapes, SYNTHIA $\to$ Cityscapes.} \label{tab5}
  \vspace{-2mm}
  \resizebox{\linewidth}{!}{
  \begin{tabular}{l | l c c c }
  \toprule[1.2pt]
  \multirow{2}{*}{Type} & \multirow{2}{*}{Method} & \multirow{2}{*}{Net.} & GT$\to$Ci & SY$\to$Ci\\
                        &                         &                       & mIoU & mIoU*  \\
  \midrule
  \multirow{3}{*}{SSDA} & MME (3.4\%)~\cite{saito2019semi} &\multirow{3}[1]{*}{V2}    & 52.6 & 59.6 \\
  & ASS (3.4\%)~\cite{wang2020differential}                 &                          & 54.2 & 62.1 \\
  & DLDM (3.4\%)~\cite{chen2021semi}                       &                          & 61.2 & 68.4 \\
  \midrule
   & \bf Ours (2.2\%) & V2 & \bf72.5 & \bf78.6 \\
  \midrule
  \midrule
  \multirow{5}{*}{SSL} & GCT (3.1\%)~\cite{ke2020guided}  &\multirow{6}[1]{*}{V3+}    & 63.2 & - \\
  & MT (3.1\%)~\cite{tarvainen2017mean}                   &                           & 64.1 & - \\
  & CCT (3.1\%)~\cite{ouali2020semi}                      &                           & 66.4 & - \\
  & AEL (3.1\%)~\cite{hu2021semi}                         &                           & 74.3 & - \\
  & U2PL+AEL (6.3\%)~\cite{chen2021semi}                  &                           & 74.9 & - \\
  \midrule
  & \bf Ours (2.2\%)     &\multirow{2}[1]{*}{V3+}   & 75.0 & 80.9 \\              
  & \bf Ours (5.0\%)     &                          & \bf76.1 & \bf82.1 \\
  \bottomrule[1.2pt]
  \end{tabular}
  } 
  \vspace{-3mm}
  \end{table}

\begin{table}[!t]
  \centering
  \caption{Experiments on \textbf{source-free domain adaptation scenario}.}\label{tab6}
  \vspace{-2mm}
  \resizebox{\linewidth}{!}{
  \begin{tabular}{l c c c | c c}
  \toprule[1.2pt]
  \multirow{2}{*}{Method} & \multirow{2}{*}{Net.} & \multirow{2}{*}{Budget} & GT$\to$Ci & \multicolumn{2}{c}{SY$\to$Ci} \\
  & & &  mIoU & mIoU & mIoU* \\
  \midrule
  URMA~\cite{fleuret2021uncertainty}  & \multirow{4}[1]{*}{V2}  & - & 45.1 & 39.6 & 45.0 \\
  LD~\cite{you2021domain}             &                         & - & 45.5 & 42.6 & 50.1 \\
  SFDA~\cite{kundu2021generalize}     &                         & - & 53.4 & 52.0 & 60.1 \\
  RIPU~\cite{xie2022towards}          &                         & 2.2\% & 67.1 & 68.7 & 74.1 \\
  \midrule
  \bf Ours                                & V2 & 2.2\% & \bf70.4 & \bf72.0 & \bf78.1 \\
  \bottomrule[1.2pt]
  \end{tabular}
  }
  \vspace{-3mm}
\end{table}

\begin{table*}[!t]
  \centering
  \caption{{The results for model upper limits.}}\label{tab7}
  \vspace{-2mm}
  \resizebox{\textwidth}{!}{
  \begin{tabular}{l | c | c | c  c  c c c c c c c c c c c c c c c c c c r | c}
  \toprule[1.2pt]
  Task & Net. & Budget &\rotatebox{60}{road} &\rotatebox{60}{side.} &\rotatebox{60}{buil.} &\rotatebox{60}{wall} &\rotatebox{60}{fence} &\rotatebox{60}{pole} &\rotatebox{60}{light} &\rotatebox{60}{sign} &\rotatebox{60}{veg.} &\rotatebox{60}{terr.} &\rotatebox{60}{sky} &\rotatebox{60}{pers.} &\rotatebox{60}{rider} &\rotatebox{60}{car} &\rotatebox{60}{truck} &\rotatebox{60}{bus} &\rotatebox{60}{train} &\rotatebox{60}{motor} &\rotatebox{60}{bike} & mIoU \\
  \midrule
  GTAV$\to$Cityscapes  & \multirow{2}[1]{*}{V3+} & 100\% & 98.0 & 84.6 & 93.0 & 58.7 & 62.9 & 67.3 & 73.4 & 80.0 & 92.8 & 63.8 & 95.1 & 84.2 & 66.1 & 95.6 & 86.6 & 91.4 & 82.9 & 70.0 & 79.3 & 80.3 \\
  SYNTHIA$\to$Cityscapes  & & 100\% & 97.9 & 84.2 & 93.1 & 63.0 & 61.8 & 68.0 & 72.9 & 81.3 & 92.9 & 66.6 & 94.9 & 84.6 & 67.0 & 95.7 & 86.8 & 92.6 & 85.6 & 71.7 & 79.4 & 81.1 \\
  \bottomrule[1.2pt]
  \end{tabular}
  }
  \vspace{-3mm}
\end{table*}
\vspace{-2mm}
\subsection{Ablation Study}
\label{sec:ablation}

To further investigate the efficacy of each component of our \method, we perform ablation studies on GTAV $\to$ Cityscapes. We randomly select 5\% of the data in the target domain to train in DeepLab-v3+ as the baseline. As shown in \Tref{tab3}, satisfactory and consistent gains from the baseline to our full method demonstrate the effectiveness of each component. Compared to the baseline, $\Lossunsup$ can leverage the unlabeled data of the target domain to improve performance by 10.38\%. The performance is further improved by 0.9\% after adding $\Losscontrast$ in \ssl. Finally, by replacing the samples in the baseline with the samples picked by active learning, our performance reaches 76.11\%, proving the effectiveness of the sample selection strategy based on prediction uncertainty.

\subsection{Further Analysis}
\label{sec:analysis_results}

\paragraph{Extension to open set domain adaptive.} SYNTHIA shares only 16 classes with Cityscapes, so previous methods only evaluate mIoU for 16 classes and 13 classes on this task. In ALDA, we will report mIoU for 19 classes on the SYNTHIA $\to$ Cityscapes task due to the addition of data from the target domain. The evaluation results are shown in \Tref{tab4}. Judging from the evaluation results of ``terrain'', ``truck'', and ``train'' missing three categories, we can still produce effects comparable to a large amount of labeled data under the premise of using a very small amount of target data.

\paragraph{Comparison with SSDA and SSL methods.} We compare our method with semi-supervised domain adaptation and semi-supervised learning methods. Among them, semi-supervised learning only uses the data of single-domain Cityscapes. The results are shown in \Tref{tab5}.

\paragraph{Extension to the source-free scenario.} Due to data privacy and constraints on computing resources, domain adaptation sometimes fails to obtain source domain datasets. We further evaluate the generalization of \method by extending to source-free domain adaptation (SFDA). Comparing methods include URMA~\cite{fleuret2021uncertainty}, LD~\cite{you2021domain}, SFDA~\cite{kundu2021generalize}, and RIPU~\cite{xie2022towards}. Results in \Tref{tab6} validate the effectiveness of \method for this challenging source-free domain adaptive task. 

\paragraph{The upper bound accuracy of the model.} To verify the robustness of the proposed method, we conduct experiments on the upper limit of the model’s accuracy, that is, introduce all target domain data for iterative learning. We adopt DeepLab-V3+ and the backbone ResNet-101 for training. As can be seen from \Tref{tab7}, \method achieves mIoUs of 80.3\% and 81.1\% in the 19-category evaluations of GTAV $\to$ Cityscapes and SYNTHIA $\to$ Cityscapes.

\paragraph{t-SNE Visualization.} We draw t-SNE visualization~\cite{van2008visualizing} of the learned feature representations for contrast methods (RIPU~\cite{xie2022towards}) and ours \method in \fref{fig5}. We randomly select an image from the target domain and then map its high-dimensional latent feature representation into 2D space. In \fref{fig5}, our method can separate the features between different categories better, and the decision boundary is clearer than other methods.

\section{Conclusion}
\label{sec:conclusion}

We propose \method to achieve the best performance in domain-adaptive semantic segmentation with minimal label cost. \method introduces semi-supervised learning in the active learning domain adaptive semantic segmentation task. By learning massive unlabeled data, the accuracy of the model in the target domain is improved, and the noise problem of pseudo-labels is corrected with limited human intervention. In addition, from the perspective of practical application, based on the existing labeling software Labelme, we propose a selection strategy based on prediction uncertainty with a single image as the smallest selection unit. The image label is corrected in combination with the prediction of the model, which further reduces the labeling cost. The effectiveness of the proposed method is verified through extensive experiments and ablation studies, and our \method achieves state-of-the-art results.

\bibliographystyle{IEEEtran}
\bibliography{IEEEexample}

\begin{thebibliography}{10}
\providecommand{\url}[1]{#1}
\csname url@samestyle\endcsname
\providecommand{\newblock}{\relax}
\providecommand{\bibinfo}[2]{#2}
\providecommand{\BIBentrySTDinterwordspacing}{\spaceskip=0pt\relax}
\providecommand{\BIBentryALTinterwordstretchfactor}{4}
\providecommand{\BIBentryALTinterwordspacing}{\spaceskip=\fontdimen2\font plus
\BIBentryALTinterwordstretchfactor\fontdimen3\font minus
  \fontdimen4\font\relax}
\providecommand{\BIBforeignlanguage}[2]{{%
\expandafter\ifx\csname l@#1\endcsname\relax
\typeout{** WARNING: IEEEtran.bst: No hyphenation pattern has been}%
\typeout{** loaded for the language `#1'. Using the pattern for}%
\typeout{** the default language instead.}%
\else
\language=\csname l@#1\endcsname
\fi
#2}}
\providecommand{\BIBdecl}{\relax}
\BIBdecl

\bibitem{9529067}
Y.~Qian, L.~Deng, T.~Li, C.~Wang, and M.~Yang, ``Gated-residual block for
  semantic segmentation using rgb-d data,'' \emph{IEEE Trans. Intell. Transp.
  Syst.}, vol.~23, no.~8, pp. 11\,836--11\,844, 2022.

\bibitem{9042876}
G.~Dong, Y.~Yan, C.~Shen, and H.~Wang, ``Real-time high-performance semantic
  image segmentation of urban street scenes,'' \emph{IEEE Trans. Intell.
  Transp. Syst.}, vol.~22, no.~6, pp. 3258--3274, 2021.

\bibitem{geiger2012we}
A.~Geiger, P.~Lenz, and R.~Urtasun, ``Are we ready for autonomous driving? the
  kitti vision benchmark suite,'' in \emph{Proc. IEEE Conf. Comput. Vis.
  Pattern Recog. (CVPR)}.\hskip 1em plus 0.5em minus 0.4em\relax IEEE, 2012,
  pp. 3354--3361.

\bibitem{cordts2016cityscapes}
M.~Cordts, M.~Omran, S.~Ramos, T.~Rehfeld, M.~Enzweiler, R.~Benenson,
  U.~Franke, S.~Roth, and B.~Schiele, ``The cityscapes dataset for semantic
  urban scene understanding,'' in \emph{Proc. IEEE Conf. Comput. Vis. Pattern
  Recog. (CVPR)}, 2016, pp. 3213--3223.

\bibitem{asgari2021deep}
S.~Asgari~Taghanaki, K.~Abhishek, J.~P. Cohen, J.~Cohen-Adad, and G.~Hamarneh,
  ``Deep semantic segmentation of natural and medical images: a review,''
  \emph{Artif. Intell. Rev.}, vol.~54, no.~1, pp. 137--178, 2021.

\bibitem{zhang2018context}
H.~Zhang, K.~Dana, J.~Shi, Z.~Zhang, X.~Wang, A.~Tyagi, and A.~Agrawal,
  ``Context encoding for semantic segmentation,'' in \emph{Proc. IEEE Conf.
  Comput. Vis. Pattern Recog. (CVPR)}, 2018, pp. 7151--7160.

\bibitem{10021219}
L.~Guan and X.~Yuan, ``Instance segmentation model evaluation and rapid
  deployment for autonomous driving using domain differences,'' \emph{IEEE
  Trans. Intell. Transp. Syst.}, pp. 1--10, 2023.

\bibitem{li2021generalized}
S.~Li, B.~Xie, Q.~Lin, C.~H. Liu, G.~Huang, and G.~Wang, ``Generalized domain
  conditioned adaptation network,'' \emph{IEEE Trans. Pattern Anal. Mach.
  Intell.}, 2021.

\bibitem{vu2019advent}
T.-H. Vu, H.~Jain, M.~Bucher, M.~Cord, and P.~P{\'e}rez, ``Advent: Adversarial
  entropy minimization for domain adaptation in semantic segmentation,'' in
  \emph{Proc. IEEE Conf. Comput. Vis. Pattern Recog. (CVPR)}, 2019, pp.
  2517--2526.

\bibitem{huo2022domain}
X.~Huo, L.~Xie, H.~Hu, W.~Zhou, H.~Li, and Q.~Tian, ``Domain-agnostic prior for
  transfer semantic segmentation,'' in \emph{Proc. IEEE Conf. Comput. Vis.
  Pattern Recog. (CVPR)}, 2022, pp. 7075--7085.

\bibitem{zhang2021transfer}
J.~Zhang, C.~Ma, K.~Yang, A.~Roitberg, K.~Peng, and R.~Stiefelhagen, ``Transfer
  beyond the field of view: Dense panoramic semantic segmentation via
  unsupervised domain adaptation,'' \emph{IEEE Trans. Intell. Transp. Syst.},
  vol.~23, no.~7, pp. 9478--9491, 2021.

\bibitem{song2020nighttime}
C.~Song, J.~Wu, L.~Zhu, M.~Zhang, and H.~Ling, ``Nighttime road scene parsing
  by unsupervised domain adaptation,'' \emph{IEEE Trans. Intell. Transp.
  Syst.}, vol.~23, no.~4, pp. 3244--3255, 2020.

\bibitem{liu2021source}
Y.~Liu, W.~Zhang, and J.~Wang, ``Source-free domain adaptation for semantic
  segmentation,'' in \emph{Proc. IEEE Conf. Comput. Vis. Pattern Recog.
  (CVPR)}, 2021, pp. 1215--1224.

\bibitem{zhang2021prototypical}
P.~Zhang, B.~Zhang, T.~Zhang, D.~Chen, Y.~Wang, and F.~Wen, ``Prototypical
  pseudo label denoising and target structure learning for domain adaptive
  semantic segmentation,'' in \emph{Proc. IEEE Conf. Comput. Vis. Pattern
  Recog. (CVPR)}, 2021, pp. 12\,414--12\,424.

\bibitem{shin2021labor}
I.~Shin, D.-J. Kim, J.~W. Cho, S.~Woo, K.~Park, and I.~S. Kweon, ``Labor:
  Labeling only if required for domain adaptive semantic segmentation,'' in
  \emph{Proc. Int. Conf. Comput. Vis. (ICCV)}, 2021, pp. 8588--8598.

\bibitem{su2020active}
J.-C. Su, Y.-H. Tsai, K.~Sohn, B.~Liu, S.~Maji, and M.~Chandraker, ``Active
  adversarial domain adaptation,'' in \emph{Proc. IEEE Winter Conf. Appl.
  Comput. Vis. (WACV)}, 2020, pp. 739--748.

\bibitem{8733203}
J.~Mei, B.~Gao, D.~Xu, W.~Yao, X.~Zhao, and H.~Zhao, ``Semantic segmentation of
  3d lidar data in dynamic scene using semi-supervised learning,'' \emph{IEEE
  Trans. Intell. Transp. Syst.}, vol.~21, no.~6, pp. 2496--2509, 2020.

\bibitem{mei2020instance}
K.~Mei, C.~Zhu, J.~Zou, and S.~Zhang, ``Instance adaptive self-training for
  unsupervised domain adaptation,'' in \emph{Proc. Eur. Conf. Comput. Vis.
  (ECCV)}.\hskip 1em plus 0.5em minus 0.4em\relax Springer, 2020, pp. 415--430.

\bibitem{zheng2021rectifying}
Z.~Zheng and Y.~Yang, ``Rectifying pseudo label learning via uncertainty
  estimation for domain adaptive semantic segmentation,'' \emph{Int. J. Comput.
  Vis. (IJCV)}, vol. 129, no.~4, pp. 1106--1120, 2021.

\bibitem{zou2018unsupervised}
Y.~Zou, Z.~Yu, B.~Kumar, and J.~Wang, ``Unsupervised domain adaptation for
  semantic segmentation via class-balanced self-training,'' in \emph{Proc. Eur.
  Conf. Comput. Vis. (ECCV)}, 2018, pp. 289--305.

\bibitem{zou2019confidence}
Y.~Zou, Z.~Yu, X.~Liu, B.~Kumar, and J.~Wang, ``Confidence regularized
  self-training,'' in \emph{Proc. Int. Conf. Comput. Vis. (ICCV)}, 2019, pp.
  5982--5991.

\bibitem{9619854}
L.~Rosas-Arias, G.~Benitez-Garcia, J.~Portillo-Portillo, J.~Olivares-Mercado,
  G.~Sanchez-Perez, and K.~Yanai, ``Fassd-net: Fast and accurate real-time
  semantic segmentation for embedded systems,'' \emph{IEEE Trans. Intell.
  Transp. Syst.}, vol.~23, no.~9, pp. 14\,349--14\,360, 2022.

\bibitem{9134735}
L.~Sun, K.~Yang, X.~Hu, W.~Hu, and K.~Wang, ``Real-time fusion network for
  rgb-d semantic segmentation incorporating unexpected obstacle detection for
  road-driving images,'' \emph{IEEE Robot. Autom. Lett.}, vol.~5, no.~4, pp.
  5558--5565, 2020.

\bibitem{li2021transferable}
S.~Li, M.~Xie, K.~Gong, C.~H. Liu, Y.~Wang, and W.~Li, ``Transferable semantic
  augmentation for domain adaptation,'' in \emph{Proc. IEEE Conf. Comput. Vis.
  Pattern Recog. (CVPR)}, 2021, pp. 11\,516--11\,525.

\bibitem{vs2021mega}
V.~Vs, V.~Gupta, P.~Oza, V.~A. Sindagi, and V.~M. Patel, ``Mega-cda: Memory
  guided attention for category-aware unsupervised domain adaptive object
  detection,'' in \emph{Proc. IEEE Conf. Comput. Vis. Pattern Recog. (CVPR)},
  2021, pp. 4516--4526.

\bibitem{wang2020differential}
Z.~Wang, M.~Yu, Y.~Wei, R.~Feris, J.~Xiong, W.-m. Hwu, T.~S. Huang, and H.~Shi,
  ``Differential treatment for stuff and things: A simple unsupervised domain
  adaptation method for semantic segmentation,'' in \emph{Proc. IEEE Conf.
  Comput. Vis. Pattern Recog. (CVPR)}, 2020, pp. 12\,635--12\,644.

\bibitem{wang2020classes}
H.~Wang, T.~Shen, W.~Zhang, L.-Y. Duan, and T.~Mei, ``Classes matter: A
  fine-grained adversarial approach to cross-domain semantic segmentation,'' in
  \emph{Proc. Eur. Conf. Comput. Vis. (ECCV)}.\hskip 1em plus 0.5em minus
  0.4em\relax Springer, 2020, pp. 642--659.

\bibitem{luo2019taking}
Y.~Luo, L.~Zheng, T.~Guan, J.~Yu, and Y.~Yang, ``Taking a closer look at domain
  shift: Category-level adversaries for semantics consistent domain
  adaptation,'' in \emph{Proc. IEEE Conf. Comput. Vis. Pattern Recog. (CVPR)},
  2019, pp. 2507--2516.

\bibitem{saito2019semi}
K.~Saito, D.~Kim, S.~Sclaroff, T.~Darrell, and K.~Saenko, ``Semi-supervised
  domain adaptation via minimax entropy,'' in \emph{Proc. Int. Conf. Comput.
  Vis. (ICCV)}, 2019, pp. 8050--8058.

\bibitem{chen2021semi}
S.~Chen, X.~Jia, J.~He, Y.~Shi, and J.~Liu, ``Semi-supervised domain adaptation
  based on dual-level domain mixing for semantic segmentation,'' in \emph{Proc.
  IEEE Conf. Comput. Vis. Pattern Recog. (CVPR)}, 2021, pp. 11\,018--11\,027.

\bibitem{ouali2020semi}
Y.~Ouali, C.~Hudelot, and M.~Tami, ``Semi-supervised semantic segmentation with
  cross-consistency training,'' in \emph{Proc. IEEE Conf. Comput. Vis. Pattern
  Recog. (CVPR)}, 2020, pp. 12\,674--12\,684.

\bibitem{hu2021semi}
H.~Hu, F.~Wei, H.~Hu, Q.~Ye, J.~Cui, and L.~Wang, ``Semi-supervised semantic
  segmentation via adaptive equalization learning,'' \emph{Proc. Adv. Neural
  Inf. Process. Syst. (NIPS)}, vol.~34, pp. 22\,106--22\,118, 2021.

\bibitem{wang2022semi}
Y.~Wang, H.~Wang, Y.~Shen, J.~Fei, W.~Li, G.~Jin, L.~Wu, R.~Zhao, and X.~Le,
  ``Semi-supervised semantic segmentation using unreliable pseudo-labels,'' in
  \emph{Proc. IEEE Conf. Comput. Vis. Pattern Recog. (CVPR)}, 2022, pp.
  4248--4257.

\bibitem{xie2020unsupervised}
Q.~Xie, Z.~Dai, E.~Hovy, T.~Luong, and Q.~Le, ``Unsupervised data augmentation
  for consistency training,'' \emph{Proc. Adv. Neural Inf. Process. Syst.
  (NIPS)}, vol.~33, pp. 6256--6268, 2020.

\bibitem{araslanov2021self}
N.~Araslanov and S.~Roth, ``Self-supervised augmentation consistency for
  adapting semantic segmentation,'' in \emph{Proc. IEEE Conf. Comput. Vis.
  Pattern Recog. (CVPR)}, 2021, pp. 15\,384--15\,394.

\bibitem{du2017robust}
B.~Du, Z.~Wang, L.~Zhang, L.~Zhang, and D.~Tao, ``Robust and discriminative
  labeling for multi-label active learning based on maximum correntropy
  criterion,'' \emph{IEEE Trans. Image Process.}, vol.~26, no.~4, pp.
  1694--1707, 2017.

\bibitem{gal2017deep}
Y.~Gal, R.~Islam, and Z.~Ghahramani, ``Deep bayesian active learning with image
  data,'' in \emph{Proc. Int. Conf. Mach. Learn. (ICML)}.\hskip 1em plus 0.5em
  minus 0.4em\relax PMLR, 2017, pp. 1183--1192.

\bibitem{fu2021transferable}
B.~Fu, Z.~Cao, J.~Wang, and M.~Long, ``Transferable query selection for active
  domain adaptation,'' in \emph{Proc. IEEE Conf. Comput. Vis. Pattern Recog.
  (CVPR)}, 2021, pp. 7272--7281.

\bibitem{prabhu2021active}
V.~Prabhu, A.~Chandrasekaran, K.~Saenko, and J.~Hoffman, ``Active domain
  adaptation via clustering uncertainty-weighted embeddings,'' in \emph{Proc.
  Int. Conf. Comput. Vis. (ICCV)}, 2021, pp. 8505--8514.

\bibitem{ning2021multi}
M.~Ning, D.~Lu, D.~Wei, C.~Bian, C.~Yuan, S.~Yu, K.~Ma, and Y.~Zheng,
  ``Multi-anchor active domain adaptation for semantic segmentation,'' in
  \emph{Proc. Int. Conf. Comput. Vis. (ICCV)}, 2021, pp. 9112--9122.

\bibitem{xie2022towards}
B.~Xie, L.~Yuan, S.~Li, C.~H. Liu, and X.~Cheng, ``Towards fewer annotations:
  Active learning via region impurity and prediction uncertainty for domain
  adaptive semantic segmentation,'' in \emph{Proc. IEEE Conf. Comput. Vis.
  Pattern Recog. (CVPR)}, 2022, pp. 8068--8078.

\bibitem{hoyer2022daformer}
L.~Hoyer, D.~Dai, and L.~Van~Gool, ``Daformer: Improving network architectures
  and training strategies for domain-adaptive semantic segmentation,'' in
  \emph{Proc. IEEE Conf. Comput. Vis. Pattern Recog. (CVPR)}, 2022, pp.
  9924--9935.

\bibitem{oord2018representation}
A.~v.~d. Oord, Y.~Li, and O.~Vinyals, ``Representation learning with
  contrastive predictive coding,'' \emph{arXiv:1807.03748. [Online]. Available:
  http://arxiv.org/abs/1807.03748}, 2018.

\bibitem{chen2017deeplab}
L.-C. Chen, G.~Papandreou, I.~Kokkinos, K.~Murphy, and A.~L. Yuille, ``Deeplab:
  Semantic image segmentation with deep convolutional nets, atrous convolution,
  and fully connected crfs,'' \emph{IEEE Trans. Pattern Anal. Mach. Intell.},
  vol.~40, no.~4, pp. 834--848, 2017.

\bibitem{chen2018encoder}
L.-C. Chen, Y.~Zhu, G.~Papandreou, F.~Schroff, and H.~Adam, ``Encoder-decoder
  with atrous separable convolution for semantic image segmentation,'' in
  \emph{Proc. Eur. Conf. Comput. Vis. (ECCV)}, 2018, pp. 801--818.

\bibitem{richter2016playing}
S.~R. Richter, V.~Vineet, S.~Roth, and V.~Koltun, ``Playing for data: Ground
  truth from computer games,'' in \emph{Proc. Eur. Conf. Comput. Vis.
  (ECCV)}.\hskip 1em plus 0.5em minus 0.4em\relax Springer, 2016, pp. 102--118.

\bibitem{ros2016synthia}
G.~Ros, L.~Sellart, J.~Materzynska, D.~Vazquez, and A.~M. Lopez, ``The synthia
  dataset: A large collection of synthetic images for semantic segmentation of
  urban scenes,'' in \emph{Proc. IEEE Conf. Comput. Vis. Pattern Recog.
  (CVPR)}, 2016, pp. 3234--3243.

\bibitem{he2016deep}
K.~He, X.~Zhang, S.~Ren, and J.~Sun, ``Deep residual learning for image
  recognition,'' in \emph{Proc. IEEE Conf. Comput. Vis. Pattern Recog. (CVPR)},
  2016, pp. 770--778.

\bibitem{deng2009imagenet}
J.~Deng, W.~Dong, R.~Socher, L.-J. Li, K.~Li, and L.~Fei-Fei, ``Imagenet: A
  large-scale hierarchical image database,'' in \emph{Proc. IEEE Conf. Comput.
  Vis. Pattern Recog. (CVPR)}.\hskip 1em plus 0.5em minus 0.4em\relax Ieee,
  2009, pp. 248--255.

\bibitem{everingham2015pascal}
M.~Everingham, S.~Eslami, L.~Van~Gool, C.~K. Williams, J.~Winn, and
  A.~Zisserman, ``The pascal visual object classes challenge: A
  retrospective,'' \emph{Int. J. Comput. Vis. (IJCV)}, vol. 111, no.~1, pp.
  98--136, 2015.

\bibitem{van2008visualizing}
L.~Van~der Maaten and G.~Hinton, ``Visualizing data using t-sne.'' \emph{J.
  Mach. Learn. Res.}, vol.~9, no.~11, 2008.

\bibitem{ke2020guided}
Z.~Ke, D.~Qiu, K.~Li, Q.~Yan, and R.~W. Lau, ``Guided collaborative training
  for pixel-wise semi-supervised learning,'' in \emph{Proc. Eur. Conf. Comput.
  Vis. (ECCV)}.\hskip 1em plus 0.5em minus 0.4em\relax Springer, 2020, pp.
  429--445.

\bibitem{tarvainen2017mean}
A.~Tarvainen and H.~Valpola, ``Mean teachers are better role models:
  Weight-averaged consistency targets improve semi-supervised deep learning
  results,'' \emph{Proc. Adv. Neural Inf. Process. Syst. (NIPS)}, vol.~30,
  2017.

\bibitem{fleuret2021uncertainty}
F.~Fleuret \emph{et~al.}, ``Uncertainty reduction for model adaptation in
  semantic segmentation,'' in \emph{Proc. IEEE Conf. Comput. Vis. Pattern
  Recog. (CVPR)}, 2021, pp. 9613--9623.

\bibitem{you2021domain}
F.~You, J.~Li, L.~Zhu, Z.~Chen, and Z.~Huang, ``Domain adaptive semantic
  segmentation without source data,'' in \emph{ACM Int. Conf. Multimedia
  (ACMMM)}, 2021, pp. 3293--3302.

\bibitem{kundu2021generalize}
J.~N. Kundu, A.~Kulkarni, A.~Singh, V.~Jampani, and R.~V. Babu, ``Generalize
  then adapt: Source-free domain adaptive semantic segmentation,'' in
  \emph{Proc. Int. Conf. Comput. Vis. (ICCV)}, 2021, pp. 7046--7056.

\end{thebibliography}

%

\begin{IEEEbiography}[{\includegraphics[width=1in,height=1.25in,clip,keepaspectratio]{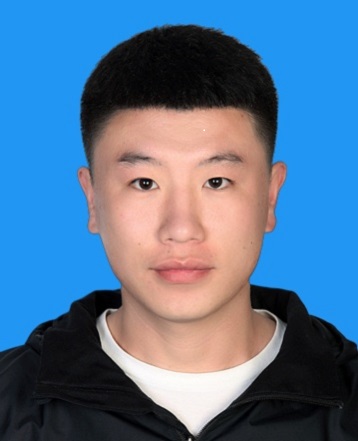}}]{Licong Guan}
received the B.S. and M.S. degrees from Qingdao University of Science and Technology, Qingdao, China, in 2017 and 2020, respectively. He is currently pursuing the Ph.D. degree with the School of Electronics and Information Engineering, Beijing Jiaotong University, Beijing, China. His research interests include image processing and pattern recognition.
\end{IEEEbiography}

\vfill

\begin{IEEEbiography}[{\includegraphics[width=1in,height=1.25in,clip,keepaspectratio]{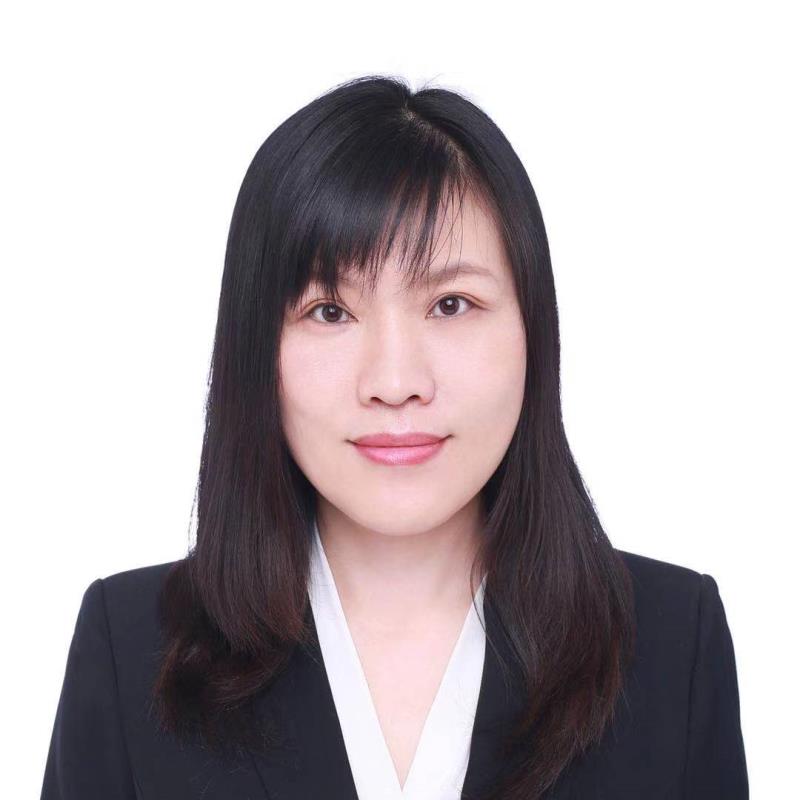}}]{Xue Yuan}
received the B.S. degree from Northeastern University, Liaoning, China, and the M.S. and Ph.D. degrees from Chiba University, Chiba, Japan, in 2004 and 2007, respectively. In 2007, she joined SECOM Company Ltd., Intelligent Systems Laboratory, Tokyo, Japan, as a Researcher. She is currently a Professor with the School of Electronics and Information Engineering, Beijing Jiaotong University, Beijing, China. Her current research interests include image processing and pattern recognition.
\end{IEEEbiography}





\end{document}